\documentclass[11pt,a4paper]{article}
\usepackage[hyperref]{naaclhlt2018}
\usepackage{times,latexsym,url}

\usepackage{booktabs}
\usepackage[english]{babel}
\usepackage{graphicx}
\usepackage{color}
\usepackage{listings}
\usepackage{setspace}
\usepackage{makecell}
\usepackage{amsmath}
\usepackage{enumitem}

\DeclareMathOperator{\loss}{loss}
\DeclareMathOperator{\scores}{scores}

\usepackage[font=small,labelfont=bf,tableposition=top]{caption}
\DeclareCaptionLabelFormat{andtable}{#1~#2,  \lstlistingname~\thelstlisting \;\& \tablename~\thetable}

\definecolor{Code}{rgb}{0,0,0}
\definecolor{Decorators}{rgb}{0.5,0.5,0.5}
\definecolor{Numbers}{rgb}{0.5,0,0}
\definecolor{MatchingBrackets}{rgb}{0.25,0.5,0.5}
\definecolor{Keywords}{rgb}{0,0,1}
\definecolor{self}{rgb}{0,0,0}
\definecolor{Strings}{rgb}{0,0.63,0}
\definecolor{Comments}{rgb}{0,0.63,1}
\definecolor{Backquotes}{rgb}{0,0,0}
\definecolor{Classname}{rgb}{0,0,0}
\definecolor{FunctionName}{rgb}{0,0,0}
\definecolor{Operators}{rgb}{0,0,0}
\definecolor{Background}{rgb}{0.98,0.98,0.98}
\lstdefinelanguage{Python}{
  numbers=left,
  numberstyle=\footnotesize,
  numbersep=1em,
  xleftmargin=1em,
  framextopmargin=2em,
  framexbottommargin=2em,
  showspaces=false,
  showtabs=false,
  showstringspaces=false,
  frame=l,
  tabsize=4,
  basicstyle=\ttfamily\small\setstretch{1},
  backgroundcolor=\color{Background},
  commentstyle=\color{Comments}\slshape,
  stringstyle=\color{Strings},
  morecomment=[s][\color{Strings}]{"""}{"""},
  morecomment=[s][\color{Strings}]{'''}{'''},
  morekeywords={import,from,class,def,for,while,if,is,in,elif,else,not,and,or,print,break,continue,return,True,False,None,access,as,,del,except,exec,finally,global,import,lambda,pass,print,raise,try,assert},
  keywordstyle={\color{Keywords}\bfseries},
  morekeywords={[2]@invariant,pylab,numpy,np,scipy},
  keywordstyle={[2]\color{Decorators}\slshape},
  emph={self},
  emphstyle={\color{self}\slshape},
}

\aclfinalcopy 

\title{UniParse: A universal graph-based parsing toolkit}

\author{Daniel Varab \and  Natalie Schluter\\
  IT University\\
  Copenhagen, Denmark \\
  {\tt \{djam,natschluter\}@itu.dk} }

\date{}
\begin{document}
\maketitle
\begin{abstract}
This paper describes the design and use of the graph-based parsing framework and toolkit UniParse, released as an open-source python software package. UniParse as a framework novelly streamlines research prototyping, development and evaluation of graph-based dependency parsing architectures.  UniParse does this by enabling highly efficient, sufficiently independent, easily readable, and easily extensible implementations for all dependency parser components. We distribute the toolkit with ready-made configurations as re-implementations of all current state-of-the-art first-order graph-based parsers, including even more efficient Cython implementations of both encoders and decoders, as well as the required specialised loss functions.
\end{abstract}

\section{Introduction}
\paragraph{Motivation.}
While graph-based dependency parsers are simple interfaces, extensible and modular implementations for sustainable parser research and development have to date been severely lacking in the research community.  Parsing research generally centres around particular components of parsers in isolation, for example, a novel decoding algorithm, the encoding of new features, a new learning algorithm, etc.  However, due to perceived gains in performance or due to the lack of foresight in writing sustainable code, these components are rarely implemented modularly or with a view to extensibility.  Both with previous sparse feature graph-based dependency parsers (such as \newcite{mcdonald_pereira2006}'s MST parser), as well as with recent state-of-the-art neural parsers (specifically \newcite{kiperwasser_goldberg2016} or \newcite{dozat_manning2017}'s neural parsers), implementations of parser components are generally hard-coupled with each other.  

Additionally, dependency parsers are often evaluated using mildly differently interpreted metrics, different data preprocessing choices, and over different target hardware. The persistently inadequate setting for parser architecture comparison entails that comparing and thereby exploring the effect of different design choices often becomes impossible to properly gauge. 




With UniParse, we provide a flexible, highly expressive, scientific framework for easy, low-barrier of entry, highly modular, highly efficient development and fair benchmarking of graph-based dependency parsing architectures.  Additionally, the framework is pre-configured with current state-of-the-art first-order sparse and neural graph-based parser implementations. 

\paragraph{Novel contributions.}
\begin{itemize}[noitemsep]\vspace{-0.4cm}
  \item We align sparse feature and neural research in graph-based dependency parsing to a \textbf{common terminology}. With this shared terminology we develop a unified framework for the UniParse toolkit to rapidly prototype new parsers and easily compare performance to previous work.  
  \item Prototyping is now rapid due to \textbf{modularity}: parser components must be developed in isolation, with no resulting loss in efficiency.  For example, measuring the empirical performance of a new decoder no longer requires implementing an encoder too, and investigating the synergy between a learning strategy and a decoder no longer requires more than a flag or calling a library function. 
  \item \textbf{Preprocessing is now made explicit} within its own component and is thereby adequately isolated and portable. 
  \item \textbf{The evaluation module is now easy to read and fully specified.}  We specify the subtle differences in computing UAS and LAS from previous literature and have implemented these in UniParse in an explicit way.  
  \item To the best of our knowledge, UniParse is the first attempt at \textbf{unifying existing dependency parsers to the same code base}. Moreover, UniParse appears to be the first attempt to enable state-of-the-art first-order sparse-feature dependency parsing within a Python environment. 
\end{itemize}\vspace{-0.3cm}
We make the parser freely available under a GNU General Public License.\footnote{\url{https://github.com/ITUnlp/UniParse}}  

A demonstration on how to easily integrate two recent, more complex, embedding components-- ELMo embeddings \cite{Peters:2018} and TCN representations \cite{Bai_etal2018}--is also made available.\footnote{\url{https://github.com/danielvarab/UniParse-extensions}}

\section{Terminology of a unified dependency parser}
Traditionally, a graph-based dependency parser consists of three components. An \emph{encoder} $\Gamma$, a set of \emph{parameters} $\lambda$, and a \emph{decoder} $h$. The possible dependency relations between all words of a sentence $S$ can modeled as a complete directed graph $G_S$ where the words are nodes and arcs are the relations.  To sub-sets of arcs from $G_S$ (called factors), $\Gamma$ associates a $d$-dimensional feature vector, its \emph{encoding}.  
The set of parameters $\lambda$ are then used to produce scores from the constructed feature vectors according to some learning architecture. These parameters are optimised over treebanks. Lastly the decoder $h$ is some maximum spanning tree algorithm with input $G_S$ and scores for factors of $G_S$ given by $\lambda$; it outputs a well-formed dependency tree, which is the raw output of a dependency model. 


Recent work on neural dependency parsers learns factor embeddings discriminatively alongside the parameters used for scoring. The result is that $\Gamma$ and $\lambda$ of dependency parsers fuse together into a union of parameters. Thus, in this work we fold the notion of encoding into the parameter space.  Now for the neural models, all parameters are trainable, whereas for sparse-feature models, the encodings of sub-sets of arcs are non-trainable.  So the unified terminology addresses only parameters $\lambda$ and a decoder $h$.




\section{API and the unified model architecture}\vspace{-0.2cm}
We provide two abstractions to implementing graph-based dependency parsers.  First, our  descriptive high-level approach focuses on expressiveness, enabling models to be described in just a few lines of code by providing an interface where the required code is minimal, only a means to configure design choices. Second, as an alternative to the high-level abstraction we emphasise that parser definition is nothing more than a composition of pre-configured low-level modular implementations. With this we invite cherry picking of the included implementations of optimised decoders, data preprocessors, evaluation module and more. We now briefly overview the basic use of the unified API and list the central low-level module implementations included with the UniParse toolkit.

\begin{figure*}[!ht]
\begin{minipage}[h]{9cm}
\includegraphics[width=9.4cm,trim={1.3cm 0 0 0},clip]{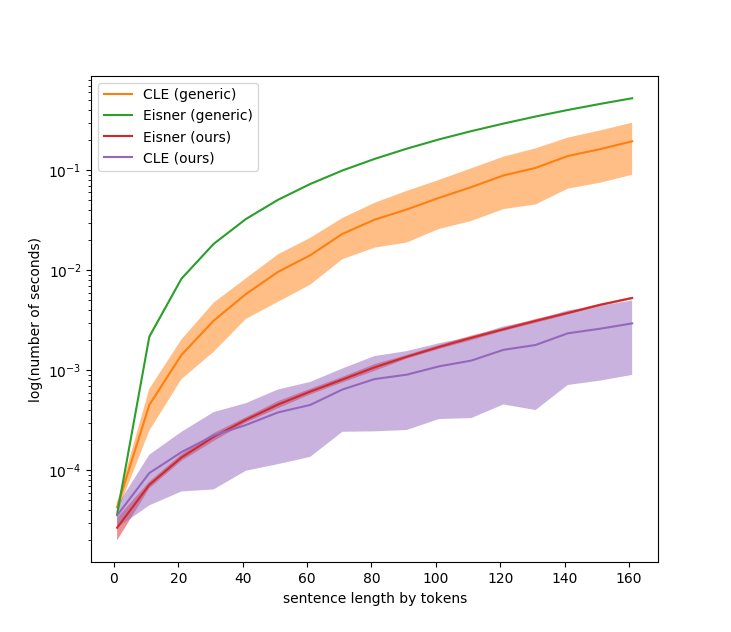}
\end{minipage}
\begin{minipage}[h]{7cm}
\begin{lstlisting}[language=Python,linewidth=7cm]{Name}
vocab = Vocabulary()
vocab = vocab.fit(train)
parser = CustomParser()
model = Model(parser, 
			  decoder="eisner",
              loss="hinge",
              optimizer="adam",
              strategy="bucket",
              vocab=vocab)
model.train(train, 
		      dev, 
              epochs=30,
              batch_size=32)
trees = model.run(test)
\end{lstlisting}
\begin{center}\begin{small}
\begin{tabular}{l|cc|c}
\toprule
\textbf{Algorithm}       & \textbf{en\_ud} & \textbf{en\_ptb} & \textbf{sents/s} \\
\midrule
Eisner (generic) & 96.35     & 479.1     & $\sim$ 80     \\
Eisner (ours)    & 1.496      & 6.31     & $\sim$ 6009   \\
CLE (generic)    & 19.12      & 93.8   & $\sim$ 404    \\
CLE (ours)       & 1.764       & 6.98     & $\sim$ 5436   \\
\bottomrule
\end{tabular}\end{small}\end{center}
\end{minipage}
\captionlistentry[table]{A table beside a figure}
\captionlistentry[lstlisting]{A table beside a figure}
\captionsetup{labelformat=andtable}
 \caption{(Right code snippet) Implementation of \newcite{kiperwasser_goldberg2016}'s neural parser in only a few lines using UniParse.  \\(Right table and left figure) Seconds a decoder takes to decode an entire dataset, given a set of scores. Score matrix entries are generated uniformly on $[0, 1]$. The random generated data has an impact on CLE since worst-case performance depends on the sorting bottleneck; the figure demonstrates this by the increasingly broad standard deviation band. Experiments are run on an Ubuntu machine with an Intel Xeon E5-2660, 2.60GHz CPU.\label{fig:kip}}
\end{figure*}

\paragraph{Elementary usage.} For ease of use we provide a high-level class to encapsulate all components of a parser.  Its use results in a significant reduction in amount of code required to implement a parser and counters unwanted boilerplate code. It provides default best-practice configurations for all included components, while enabling custom implementation whenever needed so long as it is callable and adheres to the framework's function definition of the specific component. The minimum requirements with the use of this interface are: decoder, loss function, optimiser, and batch strategy. In Figure \ref{fig:kip} is an example implementation of \newcite{kiperwasser_goldberg2016}'s neural parser in only a few lines.  The full list of possible arguments along with their interfaces can be found in the toolkit documentation. 

\paragraph{Vocabulary.} This module preprocesses a CoNLL-U formatted dataset and provides a mapper and lookup from tokens to identifiers, with support for alignment according to pre-trained word embeddings. 
Text preprocessing strategies have significant impact on NLP model performance. Despite this, little effort has put into describing such techniques in recent literature, 
which obfuscates where a model's contribution actually lies.  In the UniParse toolkit, we have included implementations for recently employed techniques within parsing for cleaning and preprocessing during the tokenisation stage.

\paragraph{Data Provider.} This module organises the tokenised data into batches for efficient learning according to the user-specified arguments. We provide several implementations for different batching strategies. This includes (1) batching by sentence length (bucketing), (2) fixed-size batching with padding, and (3) scaled padded batching through approximate clustering \cite{dozat_manning2017}.\footnote{This latter strategy is not explained in the paper but may be observed from the published TensorFlow implementation. For a description, we refer to the toolkit's README.}


\paragraph{Decoders.} We include optimised Cython implementations of first-order decoders with the toolkit (as well as Python versions of these for comparison), including both our implementations and ``generic''\footnote{Available from the Lisbon Machine Learning Summer School's public github repository \url{https://github.com/LxMLS/lxmls-toolkit/blob/1bdc382e509d24b24f581c1e1d78728c9e739169/lxmls/parsing/dependency_decoder.py}} implementations: Eisner's algorithm \cite{Eisner:1996:TNP:992628.992688} and Chu-Liu-Edmonds (CLE) \cite{chuliu_1965,edmonds_1967,zwick_2013}.  We compare Cython implementations in Table \ref{fig:kip} and Figure \ref{fig:kip} over randomised score input.  Note that our implementations are significantly faster.

\paragraph{Evaluation.}  
Unlabeled attachment score (UAS) and labeled attachment score (LAS) are central dependency parser performance metrics, measuring unlabeled and labeled arc accuracy respectively 
 with $\textrm{UAS} = \frac{\textrm{\#correct arcs}}{\textrm{\#arcs}}$ and
$\textrm{LAS} = \frac{\textrm{\#correctly labeled arcs}}{\textrm{\#arcs}}$.
Unfortunately, there are also a number unreported preprocessing choices preceding the application of these metrics, which renders direct comparison of parser performance in the literature futile, regardless of how well-motivated these preprocessing choices are.  These are generally discovered by manually screening the code implementations when these implementations are made available to the research community.  Two important variations found in state-the-art parser evaluation are the following.
\begin{table*}[!ht]\begin{center}\begin{small}
\begin{tabular}{r|l|ll|llll}\toprule
Parser configurations& \centering Dataset&
\thead{UAS n.p.\\ original }&
\thead{LAS n.p.\\ original }&
\thead{UAS \\ n.p.}&
\thead{LAS \\ n.p.}&
\thead{UAS \\ w.p.}&
\thead{LAS \\ w.p.}\\
\midrule
Kiperwasser and Goldberg   & en\_ud  & ---       & ---       & 87.71   & 84.83    & 86.80   & 85.12     \\
 (2016)                                 & en\_ptb & 93.32 & 91.2    & 93.14   & 91.57    & 92.56   & 91.17     \\
                                            & da         & ---       & ---       & 83.72   & 79.49    & 83.24   & 79.62     \\
Dozat and Manning              & en\_ud  & ---       & ---       & 91.47   & 89.38    & 90.74   & 89.01     \\
    (2017)    		                    & en\_ptb & 95.74 & 95.74  & 95.43   & 94.06    & 94.91   & 93.70     \\
       				                        & da         & ---       & ---       & 87.84   & 84.99    & 87.42   & 84.98     \\
\midrule
MSTparser   		                 & en\_ud   & ---      & ---     & 75.55   & 66.25    & 73.47   & 65.20     \\
(2006) + extensions             & en\_ptb & ---		 & ---	    & 76.07   & 64.67    & 74.00   & 63.60	   \\
 					                         & da 	      & ---		 & ---     & 68.80   & 55.30    & 67.17   & 55.52     \\
\bottomrule
\end{tabular}
\end{small}\end{center}
\caption{UAS/LAS for included parser configurations. We provide results with (w.p.) and without (n.p) punctuation. For the English universal dependencies (UD) dataset we exclude the github repository suffix \textit{EWT}. Regarding \cite{dozat_manning2017}, despite having access to the published TensorFlow code of we never observed scores exceed 95.58.  Scores for neural parsers are averages of 10 runs for the \cite{kiperwasser_goldberg2016} reimplementation and 3 runs for the \cite{dozat_manning2017} reimplementation--this difference in number of runs reflects running time of the corresponding parsers.}
\label{model-metrics}\vspace{-1cm}
\end{table*}

\begin{enumerate}[leftmargin=*,noitemsep]
\item \textbf{Punctuation removal.} Arcs incoming to any punctuation are sometimes removed.  Moreover, the definition of punctuation is not universally shared.  We provide a clear Python implementation for these metrics with and without punctuation arc deletion before application, where the definition of punctuation is clear: punctuation refers to tokens that consist of characters complying to the Unicode punctuation standard.\footnote{\url{https://www.compart.com/en/unicode/category}} This is the strategy employed by the widely used Perl evaluation script, which to our knowledge, originates from the CoNLL 2006 and 2007 shared tasks.\footnote{\url{https://depparse.uvt.nl/SoftwarePage.html\#eval07.pl}} We infer this from references in \cite{Buchholz06conll-xshared}. 
\item \textbf{Label prefixing.} Some arc labels are ``composite'', their components separated by a colon.  An example from the English Universal Dependencies data set is the label \texttt{obl:tmod}. 
The official CoNLL 2017 shared-task evaluation script\footnote{\url{http://universaldependencies.org/conll17/baseline.html}} 
allows partial matching of labels, for example matching to the non-language-specific label prefix \texttt{obl} within the language-specific label \texttt{obl:tmod} for full points. We include this variant in UniParse's evaluation module.\vspace{-0.1cm} 
\end{enumerate}

\vspace{-0.2cm}\paragraph{Callbacks.}  We include a number of useful callback utilities, such as a Tensorboard logger\footnote{\url{https://github.com/tensorflow/tensorboard}} and a patience mechanism for early stopping together with a model saver over iterations.

\vspace{-0.2cm}\paragraph{Loss Functions.}
State-of-the-art sparse feature parsers (and UniParse's included specification for this) evade direct loss computation (for example, using insights by \newcite{CrammerS03}), directly computing parameter adjustments from feature vectors. To train neural parser models we formalise a function interface and provide a set of common loss functions.  The possibilities for loss in graph-based neural dependency models have not been greatly explored.  Rather than typical loss, which is calculated over all predictions, in parsing loss has been computed over only a subset of the predicted score matrix. We impose a loss function to adhere to the type definition $\loss = f(\scores, y_p, y_g)$, where $\scores$ is the score tensor produced by the neural model, $y_p$ an optional predicted tree, and $y_g$ the gold tree. 







\vspace{-0.2cm}\paragraph{Included parser configurations.}  We include three state-of-the-art first-order dependency parser implementations as example configurations of UniParse: \newcite{mcdonald_pereira2006}'s MST sparse feature parser reimplementation\footnote{Note that this MST parser implementation consists of a restricted feature set and is only a first-order parser, as proof of concept.}, and \newcite{kiperwasser_goldberg2016} and \newcite{dozat_manning2017}'s respective graph-based neural parsers.  Experiments are carried out on English and Danish: the Penn Treebank \cite{Marcus:1994:PTA:1075812.1075835} (en\_ptb, training on sections 2-21, development on section 22 and testing on section 23), converted to dependency format following the default configuration of the Standford Dependency converter (version $>=$ 3.5.2), and the English (en\_ud), and Danish (da) datasets from version 2.1 of the universal dependencies project \cite{ud}.  Tables \ref{model-metrics} shows how our parser configurations perform compared with the originally reported parser performance.

\section{Concluding remarks}\vspace{-0.1cm}
In this paper, we have described the design and usage of UniParse, a high-level un-opinionated framework and toolkit that supports both feature-based models with on-line learning techniques, as well as recent neural architectures trained through backpropagation. We have presented the framework as answer to a long-standing need for highly efficient, easily extensible, and, most of all, directly comparable graph-based dependency parsing research.


%


%
%
    %
    %
    %
    %
    %
    %
\bibliography{parsing}
\bibliographystyle{acl_natbib}
\end{document}